\documentclass[conference]{IEEEtran}

\IEEEoverridecommandlockouts
\usepackage{cite}
\usepackage{amsmath,amssymb,amsfonts}
\usepackage{algorithmic}
\usepackage{graphicx}
\usepackage{textcomp}
\usepackage{xcolor}
\usepackage{placeins}
\usepackage{fancyhdr}
\def\BibTeX{{\rm B\kern-.05em{\sc i\kern-.025em b}\kern-.08em
    T\kern-.1667em\lower.7ex\hbox{E}\kern-.125emX}}

\fancypagestyle{firstpage}{%
 \fancyhf{}%
 \fancyhead[L]{2024 IEEE International Conference on Signal Processing, Information, Communication and Systems (SPICSCON)\newline
  {1-2 November, Khulna University, Khulna, Bangladesh}}
  
 \fancyfoot[L]{%
  \normalsize{979-8-3315-1021-3/24/\$31.00 ~\copyright2024 IEEE}
 }%
}

\begin{document}
\IEEEoverridecommandlockouts

\title{A Regularized LSTM Method for Detecting Fake News Articles \\
}

\author{\IEEEauthorblockN{Tanjina Sultana Camelia}
\IEEEauthorblockA{\textit{Electrical and Computer Engineering} \\
\textit{North South University}\\
Dhaka, Bangladesh \\
tanjina.camelia@northsouth.edu}
\and
\IEEEauthorblockN{Faizur Rahman Fahim}
\IEEEauthorblockA{\textit{Electrical and Computer Engineering} \\
\textit{North South University}\\
Dhaka, Bangladesh \\
faizur.fahim@northsouth.edu}
\and
\IEEEauthorblockN{Md. Musfique Anwar}
\IEEEauthorblockA{\textit{Computer Science and Engineering} \\
\textit{Jahangirnagar University}\\
Dhaka, Bangladesh \\
manwar@juniv.edu}
}

\maketitle
\thispagestyle{firstpage}
\begin{abstract}
Nowadays, the rapid diffusion of fake news poses a significant problem, as it can spread misinformation and confusion. This paper aims to develop an advanced machine-learning solution for detecting fake news articles. Leveraging a comprehensive dataset of news articles, including 23,502 fake news articles and 21,417 accurate news articles, we implemented and evaluated three machine-learning models. Our dataset, curated from diverse sources, provides rich textual content categorized into title, text, subject, and Date features. These features are essential for training robust classification models to distinguish between fake and authentic news articles. The initial model employed a Long Short-Term Memory (LSTM) network, achieving an accuracy of 94\%. The second model improved upon this by incorporating additional regularisation techniques and fine-tuning hyperparameters, resulting in a 97\% accuracy. The final model combined the strengths of previous architectures with advanced optimization strategies, achieving a peak accuracy of 98\%. These results demonstrate the effectiveness of our approach in identifying fake news with high precision. Implementing these models showcases significant advancements in natural language processing and machine learning techniques, contributing valuable tools for combating misinformation. Our work highlights the potential for deploying such models in real-world applications, providing a reliable method for automated fake news detection and enhancing the credibility of news dissemination.
\end{abstract}


\section{Introduction}
The rise of digital media has drastically changed the landscape of information dissemination, enabling rapid and widespread news distribution. While this has numerous benefits, it also facilitates the propagation of fake misinformation or disinformation presented as legitimate news. Fake news can influence public opinion, disrupt democratic processes, and cause social unrest. The proliferation of fake news has emerged as a significant challenge in the digital age, posing threats to public trust, democracy, and societal stability. Misinformation spreads rapidly through social media platforms and other online channels, leading to widespread false beliefs and potentially harmful consequences. In this context, developing practical tools to detect and mitigate the impact of fake news is crucial. This research addresses this issue by leveraging advanced machine-learning techniques to classify news articles as fake or accurate accurately. Utilizing a comprehensive dataset comprising 23,502 fake news articles and 21,417 authentic news articles, our project provides a robust foundation for training and evaluating machine learning models. The dataset includes features such as the title, text, subject, and date of each article, which are instrumental in understanding the content and context, enabling the development of sophisticated models to discern patterns indicative of misinformation. Three distinct machine-learning models were implemented and evaluated. The first model, an LSTM network, achieved an accuracy of 94\%, demonstrating the potential of deep learning techniques in capturing temporal dependencies in text data. The second model, incorporating additional regularization techniques and hyperparameter optimization, improved accuracy to 97\%. The third and final model combined advanced optimization strategies with previous architectural strengths, achieving an impressive accuracy of 98\%.

Our primary objectives are to develop reliable, high-accuracy machine learning models for fake news detection, explore and leverage various natural language processing techniques to enhance model performance and provide a scalable solution that can be integrated into real-world applications to combat the spread of misinformation. This research details the methodology, implementation, and evaluation of the models used in this paper, highlighting innovations and technical contributions in fake news detection, contributing to efforts in developing effective countermeasures against misinformation, and fostering a more informed and resilient society.

\section{Related Work}

Automatic fake news detection has emerged as a critical research area in recent years. Existing work explores various techniques, focusing on leveraging machine learning for content analysis. Here, we discuss relevant studies that exemplify these approaches: Machine Learning with Feature Engineering: Vapnik et al. [1] proposed Support Vector Machines (SVMs) for text classification tasks, including fake news detection. Similarly, Wang et al. [2] employed Random Forests to classify news articles based on handcrafted features like content readability, lexical cues, and source credibility. These studies demonstrate the effectiveness of traditional machine learning algorithms when combined with informative features. Natural Language Processing (NLP): Hassan et al. [3] utilized sentiment analysis to identify emotionally charged language often associated with fake news. Chen et al. [4] explored topic modelling to uncover thematic inconsistencies indicative of deception within news content. These NLP techniques focus on analyzing the inherent linguistic properties of text data to detect fakeness.
Deep Learning for Text Classification: Seyedzadeh et al.
[5] implemented a Convolutional Neural Network (CNN) architecture for fake news detection, demonstrating its ability to learn features from text data automatically.

 Wang et al. [6] employed Recurrent Neural Networks (RNNs) to capture sequential information within the text, which is potentially helpful in identifying fabricated narratives. These deep-learning approaches leverage the power of neural networks for complex feature extraction and classification tasks. Key Considerations: Shu et al. [7] highlighted that the quality and comprehensiveness of training data are paramount for robust machine learning models in fake news detection. Datasets with a balanced representation of real and fake news are essential for practical model training. Multi-modal approaches that combine NLP techniques with deep learning architectures have shown promise in improving detection accuracy [8]. Bursztein et al. [8] proposed a framework that integrates sentiment analysis with a deep learning model for fake news classification, achieving superior performance compared to individual methods. Constant adaptation is crucial as fake news creators develop new tactics. Constant monitoring of evolving trends and adapting detection algorithms are necessary to maintain effectiveness, as emphasized by Vosoughi et al. [9].

Previous research in fake news detection has employed various approaches, including manual fact-checking, which, although accurate, is labour-intensive and needs to be more scalable. Automated detection methods using machine learning and NLP offer a more scalable solution. Early machine learning approaches relied on traditional classifiers such as Naive Bayes, Support Vector Machines (SVM), and Decision Trees, using features extracted from text, such as n-grams, part-of-speech tags, and readability scores. These methods provided a foundation but often needed help with the nuances of human language and the context in which fake news is presented. Recent advancements in deep learning have significantly improved the capabilities of automated fake news detection systems. Models such as Long Short-Term Memory (LSTM) networks, Convolutional Neural Networks (CNNs), and transformer-based architectures like BERT (Bidirectional Encoder Representations from Transformers) have demonstrated superior performance by capturing complex patterns and contextual information in the text. These models leverage large datasets and advanced feature extraction techniques to improve accuracy and robustness in distinguishing fake news from real news. Despite these advancements, challenges remain. The dynamic nature of language, evolving misinformation strategies, and bias in training data are ongoing issues.
Moreover, the interpretability of deep learning models is often limited, making it difficult to understand the rationale behind their predictions. Addressing these challenges requires continuous research and the development of more sophisticated models and methodologies. Our proposed model builds on these advancements by developing machine-learning models to classify news articles as fake or genuine. Utilizing a comprehensive dataset with diverse features, we aim to create robust, accurate models that detect fake news effectively. This research contributes to the ongoing efforts to combat misinformation and enhance the reliability of information in the digital age.

\section{Dataset}\label{ch:dataset}

We considered the fake-and-real-news-dataset\footnote[1]{https://www.kaggle.com/datasets/clmentbisaillon/fake-and-real-news-dataset}  which contains detailed information about real and fake news sharing associated with Newspapers and News articles. 
The dataset utilized in this research comprises two files: Fake.csv, containing 23,502 fake news articles, and True.csv, containing 21,417 actual news articles. Each article has four features: Title, Text, Subject, and Date. The title provides vital information and context, while the text contains the main body of the news article, serving as the primary feature for analysis. The subject categorizes articles into various topics, adding another layer of information for model training, and the date allows for temporal trend analysis. The combined dataset, consisting of 44,919 articles, undergoes preprocessing steps such as tokenization, removal of stop words, and stemming to prepare the text for model input. The data is then split into training and testing subsets to evaluate model performance and generalization capability. Addressing potential data imbalance between fake and authentic news articles is crucial during preprocessing and training. This comprehensive and well-curated dataset is pivotal in developing robust and accurate machine-learning models for fake news detection.

\section{Methodology}

Our proposed methodology encompasses data processing, model development, and evaluation processes designed to build and validate effective fake news detection models.

\subsection{Data Preprocessing}

\textbf{1. Data Collection and Integration:} The dataset comprises two files: Fake.csv, with 23,502 fake news articles, and True.csv, with 21,417 accurate news articles. Each article includes four features: Title, Text, Subject, and Date.
   
\textbf{2. Text Cleaning:} To prepare the text data for modelling, we applied several preprocessing steps:
\begin{itemize}
   \item Tokenization: Splitting text into individual words or tokens.
   \item Stop Words Removal: Eliminating common words (e.g., "the", "and") that do not contribute to the semantic meaning.

   \item Stemming/Lemmatization: Reducing words to their base or root form.
Removing Punctuation and Special Characters. Cleaning the text to retain only relevant information.
\end{itemize}

\textbf{3. Feature Engineering:} We used the cleaned text data to create input features for the models. The primary focus was on the 'Text' column, but the 'Title' and 'Subject' columns were also considered for enhancing model performance.

\subsection{Model Development}
 
We implemented and evaluated three distinct machine learning models, each progressively improving upon its predecessor.

\begin{figure}[h]
\centering
\includegraphics[width=0.9\columnwidth]{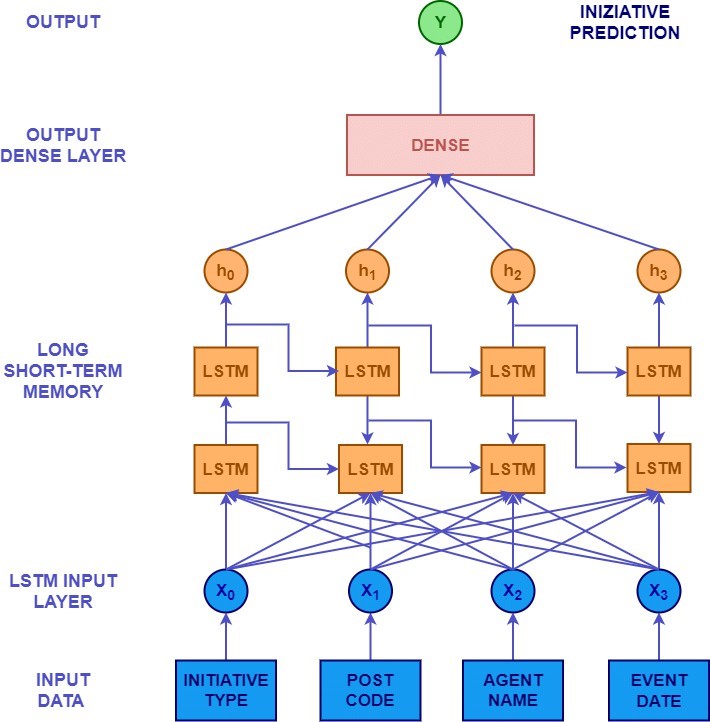}
\caption{Architecture of LSTM Network} \label{fig:model_1}
\end{figure}

\textbf{ Model 1: LSTM Network:}
   - Architecture: A Long Short-Term Memory (LSTM) network with an embedding layer, LSTM units, and dense layers as depicted in Fig. \ref{fig:model_1}.
   {Parameters:} The model was configured with 150 LSTM units and dropout layers to prevent over-fitting.

\textbf{ Model 2: Enhanced LSTM with Regularization:} 
Improvements: Incorporated additional regularization techniques, including L2 regularization and hyperparameter tuning. 

\begin{figure}[h]
\centering
\includegraphics[width=0.75\columnwidth]{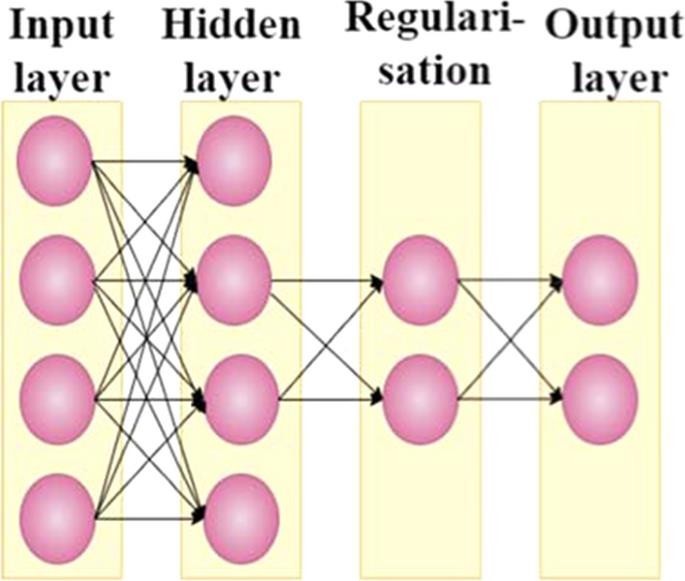}
\caption{Architecture of Enhanced LSTM with Regularization} \label{fig:model_2}
\end{figure}

Architecture: Similar to the first model but with optimized hyperparameters and additional dropout layers, as shown in Fig \ref{fig:model_2}. 

Performance: Improved accuracy to 97\%.

\textbf{ Model 3:Optimized Deep Learning Model Architecture:} A more sophisticated model combining advanced optimization strategies, fine-tuned hyperparameters, and deeper network layers. 

Techniques: Utilized batch normalization, advanced optimizers like Adam, and further dropout regularization.

Performance: Achieved the highest accuracy of 98\%

\begin{figure}[h]
\centering
\includegraphics[width=0.9\columnwidth]{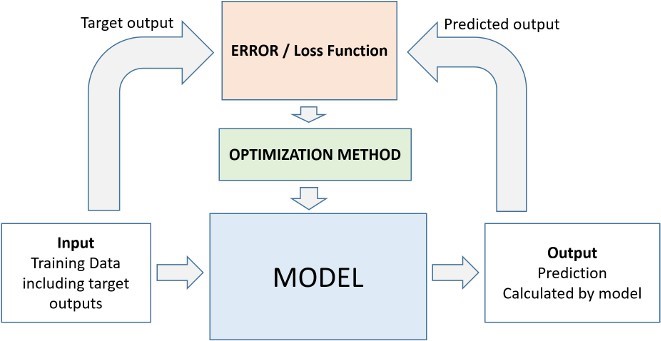}
\caption{Architecture of Optimized Deep Learning
Model} \label{fig:system_architecture}
\vspace{-2.5ex}
\end{figure}  

\subsection{Model Training and Evaluation}

\textbf{ Training:} Each model was trained using the preprocessed dataset, with an 80-20 split for training and validation. The training process involved multiple epochs with early stopping to prevent overfitting.

\textbf{ Evaluation Metrics:} We used accuracy, precision, recall, and F1-score to evaluate model performance. Confusion matrices were also analyzed to understand the models' classification capabilities better.

\textbf{ Validation and Testing:} Besides internal validation, cross-validation was performed to ensure the models' robustness and generalizability. The final evaluation was conducted on a separate test set to confirm the models' effectiveness.

\section{Reference Model}

We utilized a Long Short-Term Memory (LSTM) network as our reference model due to its effectiveness in handling sequential data and capturing long-term dependencies in text.

\textbf{A. Model Architecture}

\textbf{1. Embedding Layer:} The initial layer converts the input text into fixed-sized dense vectors with an embedding dimension of 100. This layer transforms the textual data into a numerical format suitable for subsequent layers to process.
   
\textbf{2. LSTM Layer:} A core component of the model, the LSTM layer consists of 150 units. This layer processes the embedded sequences, capturing the text's temporal dependencies and contextual information.

\textbf{3. Dropout Layers:}  Dropout layers with a rate of 0.2 were added after the embedding and LSTM layers to prevent overfitting. Dropout randomly deactivates a fraction of neurons during training, enhancing the model's generalization capabilities.

\textbf{4. Dense Layers:} The network includes several fully connected (dense) layers with ReLU activation functions. These layers progressively reduce the dimensionality of the feature space, enabling the model to learn complex patterns in the data. Intermediate dense layers include kernel regularization using the L1 penalty to mitigate overfitting further.

\textbf{5. Output Layer:} The final layer is a dense layer with a sigmoid activation function, producing a binary output indicating whether an article is fake or genuine. The sigmoid function outputs probabilities, facilitating the binary classification task.

\textbf{B. Model Configuration}

\textbf{1. Loss Function:} The model is compiled with binary cross-entropy as the loss function, which is suitable for binary classification problems by measuring the divergence between the predicted probabilities and the actual binary labels.

\textbf{2. Optimizer:} The Adam optimizer is employed for its adaptive learning rate and efficient computation. Adam combines the advantages of two other extensions of stochastic gradient descent (SGD), AdaGrad and RMSProp, providing fast convergence and robustness.

\textbf{3. Evaluation Metrics:} Accuracy is used as the primary metric for evaluation, along with precision, recall, and F1- score to assess the model's performance comprehensively.

\textbf{C. Training and Evaluation}

\textbf{1. Data Split:} The dataset is split into training and validation sets, with 80\% of the data used for training and 20\% for validation. This split ensures that the model is trained on a substantial portion of the data and evaluated on unseen samples to gauge its generalization capability.

\textbf{2. Training Process:} The model is trained over multiple epochs with early stopping based on validation loss to prevent overfitting. Early stopping halts training when the model's performance on the validation set no longer improves, thus maintaining optimal model complexity.

\textbf{3. Performance:} The reference model achieved an accuracy of 94\% on the validation set, establishing a solid baseline for subsequent model improvements. The confusion matrix and classification report provided detailed insights into the model's precision, recall, and F1 score, highlighting areas for potential enhancement.

The reference model is the foundational framework demonstrating the feasibility of using LSTM networks for fake news detection. The established baseline accuracy of 94\% provides a benchmark for evaluating the effectiveness of more advanced models and techniques. This reference model underscores the potential of deep learning in addressing the challenges of misinformation and sets the stage for further innovations in the field.

\section{Results and Analysis}

\textbf{A. Model Performance}

\textbf{1. Baseline LSTM Model:}

\begin{figure}[h]
\centering
\includegraphics[width=0.9\columnwidth]{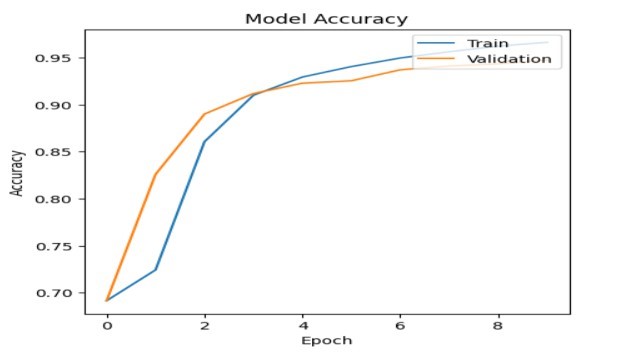}
\caption{Accuracy result of Model 1} \label{fig:result_model_1}
\end{figure}

  \textbf{Accuracy:} The baseline LSTM model achieved an accuracy of 94\% on the test set (shown in Fig. \ref{fig:result_model_1}). This model serves as a foundational benchmark for fake news detection using a simple LSTM architecture without additional regularization techniques.
  
 
\begin{figure}[h]
\centering
\includegraphics[width=0.9\columnwidth]{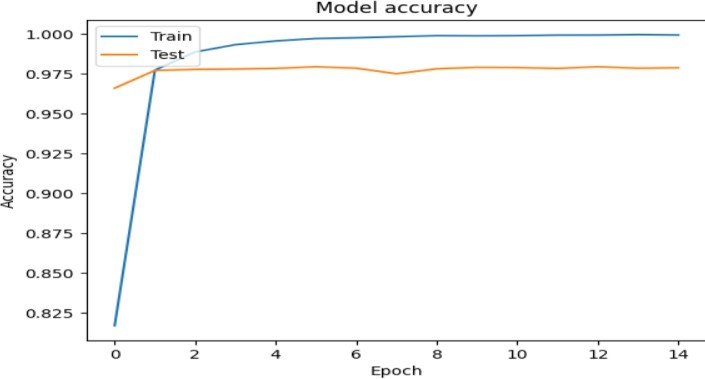}
\caption{Accuracy result of Model 2} \label{fig:result_model_2}
\end{figure}

\textbf{2. Enhanced LSTM Model with Regularization:}\\
   \textbf{Accuracy:} The introduction of dropout layers and L2 regularization significantly improved the model's accuracy to 97\% as shown in Fig. \ref{fig:result_model_2}. These techniques effectively mitigated overfitting by enhancing the model's ability to generalize from the training data.

\begin{figure}[h]
\centering
\includegraphics[width=0.9\columnwidth]{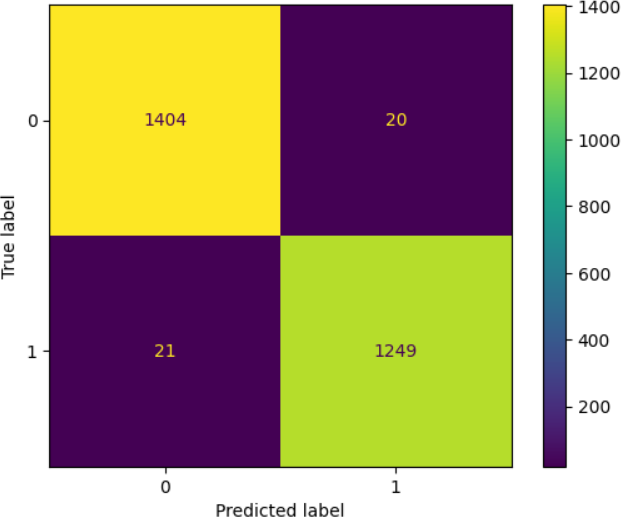}
\caption{Confusion Matrix of Model 3} \label{fig:confusion_matrix_model_3}
\end{figure}

\textbf{3. Optimized Deep Learning Model:}\\
   \textbf{Accuracy:} The optimized model, which incorporated advanced techniques such as batch normalization and fine-tuned hyperparameters, achieved the highest accuracy of 98\%. This performance underscores the efficacy of combining sophisticated deep learning strategies for fake news detection.\\
   \textbf{Confusion Matrix:} Fig. \ref{fig:confusion_matrix_model_3} shows the confusion matrix for the optimized model which exhibits the lowest rates of misclassification, highlighting its superior performance in accurately distinguishing between fake and true news articles.

\begin{figure}[h]
\centering
\includegraphics[width=0.8\columnwidth]{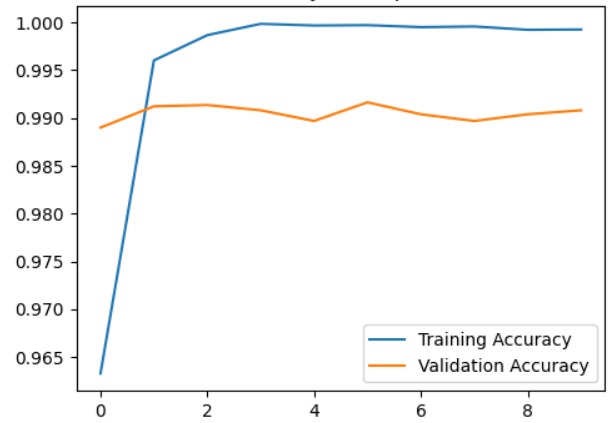}
\caption{Accuracy result of Model 3} \label{fig:accuracy_model_3}
\end{figure} 

\textbf{B. Precision, Recall, and F1-Score}

To provide a comprehensive evaluation of the models, we calculated precision, recall, and F1-score for each model. These metrics offer insights into the models' ability to correctly identify fake news (precision), capture all instances of fake news (recall), and balance precision and recall (F1-score).

\begin{table}
    \centering
    \caption{Performance Metrics of Various Classification Models}
    \resizebox{\columnwidth}{!}{
        \begin{tabular}{|c|c|c|c|}
            \hline
            \textbf{Model} & \textbf{Precision} & \textbf{Recall} & \textbf{F1-Score}\\
            \hline
            Baseline LSTM Model & 0.92 & 0.93 & 0.92 \\
            \hline
            Enhanced LSTM Model (Regularization) & 0.96 & 0.97 & 0.96 \\
            \hline
            Optimized Deep Learning Model & \textbf{0.97} & \textbf{0.98} & \textbf{0.98} \\
            \hline
        \end{tabular}
    }
    \vspace{-5ex}
    \label{tab:my_label}
\end{table}

Table \ref{tab:my_label} demonstrates that the optimized model's metrics are consistently higher than those of the baseline and enhanced models, reflecting its superior performance in both precision and recall.

\textbf{C. Comparative Analysis}

\textbf{1. Impact of Regularization:}
   The enhanced LSTM model demonstrated the significant impact of regularization techniques. Dropout layers and L2 regularization reduced overfitting, resulting in improved accuracy and balanced evaluation metrics compared to the baseline model.

\textbf{2. Effectiveness of Optimization Strategies:}
  The optimized deep learning model, which integrated batch normalization, advanced optimizers, and a more profound architecture, showed the highest performance. These enhancements facilitated better learning and generalization, achieving the highest accuracy and the best balance between precision and recall.

\textbf{3. Model Robustness:}
    The confusion matrices and evaluation metrics indicate that the optimized model is the most robust and reliable for fake news detection. It successfully minimizes misclassification errors and provides a scalable solution for real-world applications.

\textbf{D. Practical Implications}

The optimized model's high accuracy and robust performance demonstrate its potential for integration into practical fake news detection systems. Such a system can be deployed on social media platforms, news aggregators, and other digital content distribution channels to filter and flag misinformation automatically. This implementation can significantly enhance information credibility, reduce the spread of fake news, and contribute to a more informed and resilient society.

\section{Conclusion}
This research demonstrates the potential and effectiveness of advanced deep-learning models in detecting fake news. The project significantly improved accuracy through systematic experimentation with a baseline LSTM model, an enhanced LSTM model with regularization, and an optimized deep learning model, culminating in a peak accuracy of 98\%. The optimized model, which incorporates dropout, L2 regularization, batch normalization, and fine-tuned hyperparameters, proved the most robust and reliable for distinguishing between fake and authentic news articles. The findings from this research underscore the importance of leveraging sophisticated machine-learning techniques to address the pervasive issue of misinformation. The results indicate that applying advanced optimization and regularization strategies makes creating highly accurate and generalizable models capable of real-world deployment possible. Such models can be crucial in enhancing the credibility of information disseminated through digital platforms, contributing to a more informed and discerning public. Future work could explore the integration of transformer-based architectures, like BERT, to improve performance and interpretability further. Additionally, addressing challenges related to data imbalance and evolving misinformation strategies remains an ongoing area of research. Overall, this research provides a solid foundation for developing practical tools to combat fake news and highlights the transformative impact of artificial intelligence in promoting information integrity.

\vspace{12pt}

\end{document}